\documentclass[conference,a4paper]{APSIPA2021}
\usepackage{amsmath}
\usepackage{graphicx}
\usepackage{multirow}
\usepackage{threeparttable}
\usepackage[backend=biber,style=ieee,]{biblatex}
\usepackage{geometry}
\usepackage{fancyhdr}
\usepackage{amsmath,amsfonts}
\usepackage{algorithmic}
\usepackage{caption}
\usepackage{multirow}
\fancypagestyle{firststyle}{
  \fancyhf{}
  \fancyhead[C]{2023 Asia Pacific Signal and Information Processing Association Annual Summit and Conference (APSIPA ASC)}
}

\begin{document}

\title{Domain Adaptation for Efficiently Fine-tuning Vision Transformer with Encrypted Images}

\author{
\authorblockN{
Teru Nagamori\authorrefmark{1},
Sayaka Shiota\authorrefmark{1} and
Hitoshi Kiya\authorrefmark{1}
}

\authorblockA{
\authorrefmark{1}
Tokyo Metropolitan University, Japan \\
E-mail: nagamori-teru@ed.tmu.ac.jp, sayaka@tmu.ac.jp, kiya@tmu.ac.jp}
}
\maketitle
\thispagestyle{firststyle}
\pagestyle{fancy}

\begin{abstract}
In recent years, deep neural networks (DNNs) trained with transformed data have been applied to various applications such as privacy-preserving learning, access control, and adversarial defenses. However, the use of transformed data decreases the performance of models. Accordingly, in this paper, we propose a novel method for fine-tuning models with transformed images under the use of the vision transformer (ViT). The proposed domain adaptation method does not cause the accuracy degradation of models, and it is carried out on the basis of the embedding structure of ViT. In experiments, we confirmed that the proposed method prevents accuracy degradation even when using encrypted images with the CIFAR-10 and CIFAR-100 datasets.
\end{abstract}

\section{Introduction}
Deep neural networks (DNNs) have been deployed in many applications including security-critical ones such as biometric authentication and medical image analysis. In addition, training a deep learning model requires a huge amount of data and fast computing resources, so cloud environments are increasingly used in various applications of DNN models. However, since cloud providers are not always reliable in general, privacy-preserving deep learning has become an urgent problem \cite{kiya2022overview, Encryption-Then-Compression}.

One of the privacy-preserving solutions for DNNs is to use encrypted images to protect visual information on images for training models and testing. In this approach, images are transformed by using a learnable encryption method, and encrypted images are used as training and testing data. The use of models trained with encrypted images enables us to use state-of-the-art learning algorithms without any modification. However, all conventional learnable encryption methods \cite{LE, Pixel-Based, madono2020, maung_privacy, qi-san, qi2023colorneuracrypt} have the problem that is the performance degradation of encrypted models compared to models without encryption.

Traditional cryptographic methods such as homomorphic encryption \cite{Homomorphic1, Homomorphic2} are one of the other privacy-preserving approaches, but the computation and memory costs are expensive, and it is not easy to apply these methods to state-of-the-art DNNs directly. Federated learning \cite{FL} allows users to train a global model without centralizing the training data on one machine, but it cannot protect privacy during inference for test data when a model is deployed in an untrusted cloud server.

Accordingly, we focus on the problem of degraded model performance when models are trained with encrypted images. In this paper, we consider reducing the performance degradation of encrypted models when using encrypted images in the vision transformer (ViT) \cite{ViT} which has high performance in image classification, and we propose a domain adaptation method to reduce the influence of encryption. 
In experiments, the proposed method is demonstrated to maintain almost the same performance as models trained with plain images in terms of image classification accuracy even when using encrypted images.

\begin{figure*}[tb]
    \centering
    \includegraphics[bb=0 0 1013 743,scale=0.35]{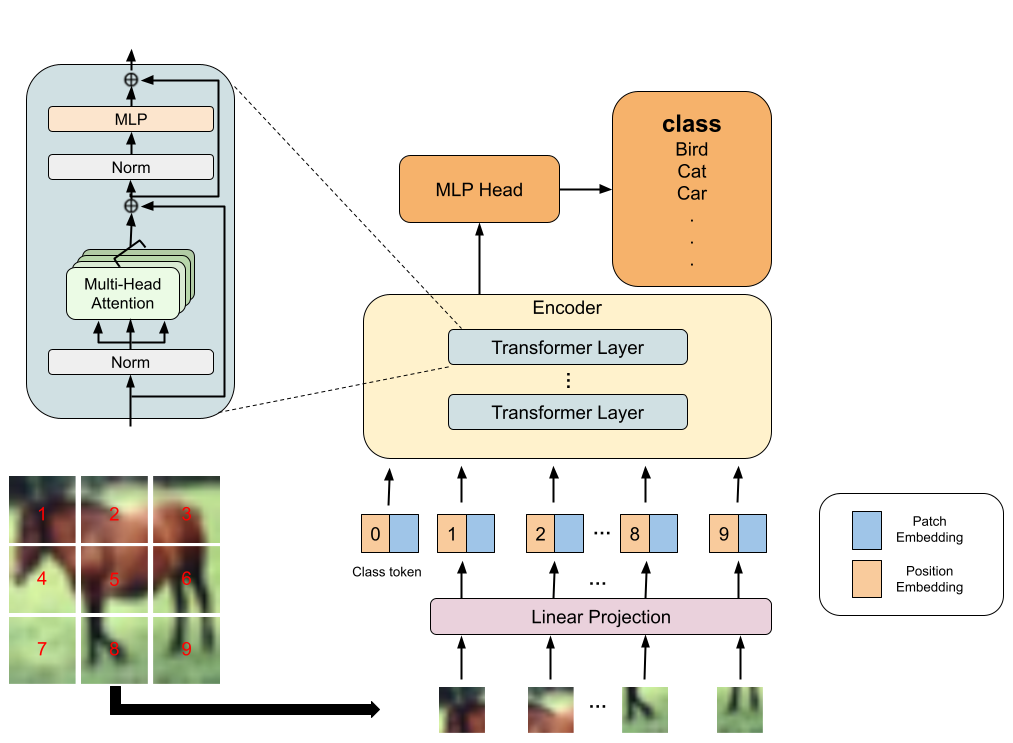}
    \caption{Architecture of Vision Transformer}
    \label{ViT}
\end{figure*}

\section{Related work}
\subsection{Image Encryption for Deep Learning}
Various image transformation methods with a secret key, often referred to as perceptual image encryption or image cryptography, have been studied so far for many applications \cite{kiya2022overview}. In this paper, we focus on learnable images transformed with a secret key, which have been studied for deep learning. Learnable encryption enables us to directly apply encrypted data to a model as training and testing data. Encrypted images have no visual information on plain images in general, so privacy-preserving learning can be carried out by using visually protected images. In addition, the use of a secret key allows us to embed unique features controlled with the key into images. Adversarial defenses \cite{maung_AD} and access control \cite{KIYA20232022} are carried out with encrypted data using unique features.

Tanaka first introduced a block-wise learnable image encryption method (LE) with an adaptation layer \cite{LE}, which is used prior to a classifier to reduce the influence of image encryption. Another encryption method is a pixel-wise encryption (PE) method in which negative-positive transformation (NP) and color component shuffling are applied without using any adaptation layer \cite{Pixel-Based}. However, both encryption methods are not robust enough against ciphertext-only attacks, as reported in \cite{chang2020attacks, Ito_access}. To enhance the security of encryption, LE was extended to ELE by adding a block scrambling step and a pixel encryption operation with multiple keys \cite{madono2020}. However, ELE still has an inferior accuracy compared with using plain images, although an additional adaptation network (denoted as ELE-AdaptNet hereinafter) is applied to reduce the influence of the encryption. Moreover, images large in size cannot be applied to ELE because of the high computation cost of ELE-AdaptNet.

Recently, some methods have been proposed to use these block-wise encryptions combined with isotropic networks such as ViT to prevent performance degradation \cite{maung_privacy, qi-san}, but this problem has not been completely solved. Another encryption method, a neural network-based image encryption method called NeuraCrypt \cite{yala2021neuracrypt}, has been proposed, but it has problems such as difficulty in using it for color images and difficulty with existing models because the input is not an image. To solve the problem, Color-NeuraCrypt \cite{qi2023colorneuracrypt}, which is an extension of NeuraCrypt, was proposed, but this method also has the same performance degradation problem as conventional block-wise encryption methods. 

Accordingly, we focus on the performance degradation problem when combining block-wise encryption and an isotropic network, and we propose a domain adaptation method to solve the problem.

\subsection{Vision Transformer}
The Vision Transformer (ViT) \cite{ViT} is commonly used in image classification tasks and is known to provide a high classification performance. As shown in Fig. \ref{ViT}, in ViT, an input image is divided into patches, and each patch is transformed into a one-dimensional trainable vector. Afterward, they are input into the transformer encoder. The encoder outputs only the class token, which is a vector of condensed information about the entire image, and it is used for classification. Also, the architecture of ViT has two types of embedding: positional embedding, which stores the location information of cropped patches in an image, and patch embedding, which transforms each patch into a trainable vector.

Previous studies have indicated that when DNN models are trained with encrypted images, the performance of the models is degraded compared to models trained with plain images. In contrast, it has been pointed out that block-wise encryption has a high affinity with models that have embedding structures such as ViT \cite{maung_privacy,jimaging8090233}. Accordingly, we focus on the affinity between block-wise image encryption and the embedding structure of ViT to reduce the influence of image encryption even when encrypted images are used for training models.

\begin{figure*}[tb]
    \centering
    \includegraphics[bb=0 0 1500 400,scale=0.48]{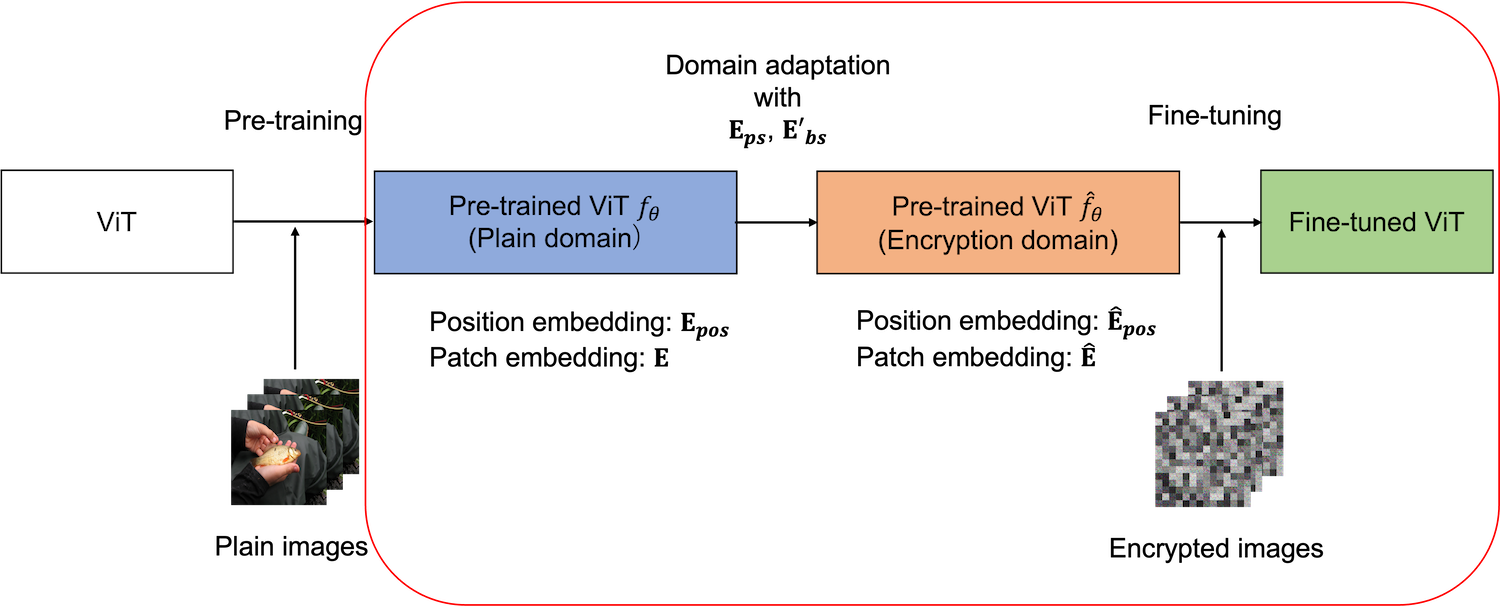}
    \caption{Fine-tuning procedure using image encryption and domain adaptation}
    \label{Overview of DA}
\end{figure*}

\section{Proposed Method}
\subsection{Overview}
Figure \ref{Overview of DA} shows the procedure for fine-tuning ViT models with encrypted images and the proposed domain adaptation. Since the pre-trained model as in \cite{ViT} was prepared in advance, the area surrounded by a red box is discussed in this paper.

In ViT, an input image $x \in \mathbb{R}^{h \times w \times c}$ is divided into $N$ patches with a size of $p \times p$, where $h$, $w$ and $c$ are the height, width, and number of channels of the image. Also, $N$ is given as $hw/p^2$. Afterward, each patch is flattened into $x_{p}^i = [x_{p}^i(1), x_{p}^i(2),\dots , x_{p}^i(L)]$. Finally, a sequence of embedded patches is given as
\begin{align}
    z_{0} =& [x_{class}; x_{p}^1\mathbf{E}; x_{p}^2\mathbf{E}; \dots x_{p}^i\mathbf{E}; \dots x_{p}^N\mathbf{E}] + \mathbf{E_{pos}},
\end{align}
where 
\begin{align*}
\mathbf{E_{pos}} =& ((e_{pos}^0)^\top (e_{pos}^1)^\top \dots (e_{pos}^i)^\top \dots (e_{pos}^N)^\top)^\top,\\
  L =& p^{2}c, x_{class} \in \mathbb{R}^{D}, \ x_{p}^{i} \in \mathbb{R}^{L}, \ e_{pos}^i \in \mathbb{R}^{D},\\
  \mathbf{E} \in& \mathbb{R}^{L \times D}, \ \mathbf{E_{pos}} \in \mathbb{R}^{(N+1) \times D}.
\end{align*}
$x_{class}$ is a class token, $\mathbf{E}$ is an embedding (patch embedding) to linearly map each patch to dimensions $D$, $\mathbf{E_{pos}}$ is an embedding (position embedding) that gives position information to patches in the image, $e_{pos}^0$ is position information of a class token, $e_{pos}^i$ is position information of each patch, and $z_0$ is a sequence of embedded patches.

In this paper, we use block scrambling and pixel shuffling used in ELE for image encryption \cite{madono2020}, where two keys are commonly applied to all images and blocks, respectively. In block scrambling, since an image is divided into non-overlapped blocks (patches) and the divided blocks are randomly permutated with a key, position embedding $\mathbf{E_{pos}}$ is affected by the position scrambling in general. In contrast, since pixel shuffling randomly permutates the position of pixels in each block, patch embedding $\mathbf{E}$ is affected by the pixel shuffling. As a result, models fine-tuned with encrypted images have a lower performance than models fine-tuned with plain images due to the influence of encryption. In other words, since pre-training a model is carried out by using plain images, the domain of fine-tuned models is different from that of pre-trained ones.

Accordingly, to reduce the influence of encryption, we propose a domain adaptation method, which is carried out on the basis of the embedding structure of ViT, for efficiently fine-tuning models.

\begin{figure*}[tb]
    \centering
    \includegraphics[bb=0 0 1498 273,scale=0.37]{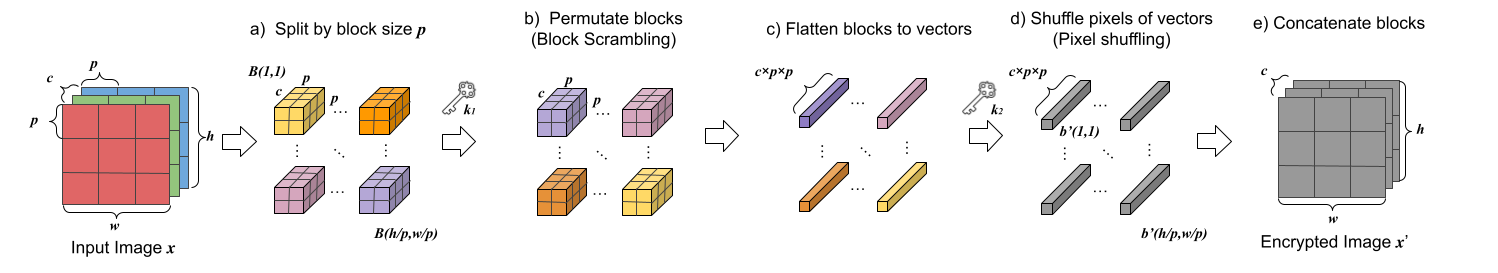}
    \caption{Procedure of image encryption}
    \label{Image Encryption}
\end{figure*}

\begin{figure*}[tb]
    \centering
    \scalebox{0.6}[0.6]{
    \begin{tabular}{cccc}
        \begin{minipage}[b]{0.4\hsize}
          \centering
          \includegraphics[bb=0 0 224 224,scale=0.6]{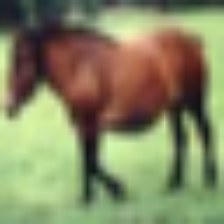}
        \end{minipage}
        &
        \begin{minipage}[b]{0.4\hsize}
          \centering
          \includegraphics[bb=0 0 224 224,scale=0.6]{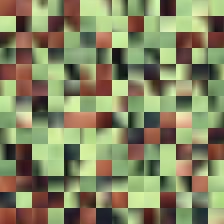}
        \end{minipage}
        &
        \begin{minipage}[b]{0.4\hsize}
          \centering
          \includegraphics[bb=0 0 224 224,scale=0.6]{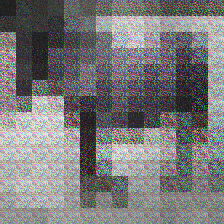}
        \end{minipage}
        &
        \begin{minipage}[b]{0.4\hsize}
          \centering
          \includegraphics[bb=0 0 224 224,scale=0.6]{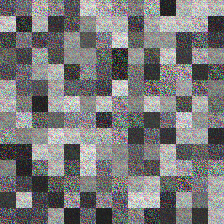}
        \end{minipage} \\
        
        \begin{tabular}{c}
            (a) Original \\ (224 $\times$ 224 $\times$ 3)
        \end{tabular}
        &
        \begin{tabular}{c}
            (b) Block scrambling \\ (block size = 16)
        \end{tabular}
        &
        \begin{tabular}{c}
            (c) Pixel shuffling \\ (block size = 16)
        \end{tabular}
        &
        \begin{tabular}{c}
            (d) Block scrambling + Pixel shuffling \\ (block size = 16)
        \end{tabular}
    \end{tabular}
    }
    \caption{Example of encrypted images}
    \label{Encrypted Images}
\end{figure*}

\subsection{Image Encryption}
Encrypted images are used to fine-tune a ViT model $f_{\theta}$ as shown in Fig. \ref{Overview of DA}.
Figure \ref{Image Encryption} illustrates the encryption procedure used in this paper. This procedure can be expressed as follows.
\begin{itemize}

    \item[(a)] Divide an image $x \in \mathbb{R}^{h \times w \times c}$ into non-overlapped blocks with a size of $p \times p$ such that $B =  \left\{B_{1},\dots ,B_{N}\right\}$ .
    
    \item[(b)] Permutate blocks with secret key $k_1$ such as
        \begin{equation} \label{eq: bs}
                    \hat{B} = \mathbf{E_{bs}}B^\top, \ \hat{B} \in \mathbb{R}^{N},
        \end{equation}

        where $\mathbf{E_{bs}}$ is generated using the key $k_1$ as follows.
        \begin{itemize}
            \item[1)]Using key $k_1$, generate a random integer vector with a length of $N$ as 
                \begin{equation}
                    l_{e} = [l_{e}(1), l_{e}(2), \dots, l_{e}(i)\ , \dots ,l_{e}(N)] \ ,
                \end{equation}
                where
                \begin{align*}
                    l_{e}(i) \in& \left\{1,2,...,N\right\}, \\
                    l_{e}(i) \neq& l_{e}(j) \ \text{if} \ i \neq j \\
                    i,j \in& \left\{1, \dots, N\right\}.
                \end{align*}
            \item[2)] Given $k_{(i,j)}$ as
                \begin{equation}
                    k_{(i,j)} = \left\{ \begin{matrix}0&(j\neq l_{e}(i))\\ 1&(j = l_{e}(i)) \end{matrix} \right. .
                \end{equation}
            \item[3)] Define $\mathbf{E_{bs}}$ as
                \begin{align} \label{eq: ebs}
                    \mathbf{E_{bs}} =
                    \begin{bmatrix} 
                          k_{(1,1)} & k_{(1,2)} & \dots & k_{(1,N)} \\
                          k_{(2,1)} & k_{(2,2)} &\dots  & k_{(2,N)} \\
                          \vdots & \vdots & \ddots & \vdots \\
                          k_{(N,1)} & k_{(N,2)} & \dots & k_{(N,N)}\\
                    \end{bmatrix},
                \end{align}
                \begin{align*}
                   \mathbf{E_{bs}} &\in \mathbb{R}^{N \times N}.
                \end{align*}
        \end{itemize}
    
    \item[(c)]Flatten each block $\hat{B_{i}}$ into a vector $\hat{b}_{i}$ such that
        \begin{equation}
            \hat{b}_{i} = [\hat{b}_{i}(1), \dots , \hat{b}_{i}(L)].
        \end{equation}

    where $\hat{b}_{i}$ is the same as $x_{p}^i$ in Eq. (1).
    
    \item[(d)]Shuffle pixels in $\hat{b}_{i}$ with secret key $k_2$  such as
        \begin{equation}
                    b'_{i} = \mathbf{E}_{ps}\hat{b}_{i}^\top, \ 
                    b'_{i} \in \mathbb{R}^{L},
        \end{equation}
        where $\mathbf{E_{ps}}$ is a matrix generated by the same procedure as $\mathbf{E_{bs}}$ ($N$ is replaced with $L$).
        
        Also, Eq. (7) is expressed as
        \begin{equation} \label{eq: ps}
                    b'_{i} = \mathbf{E}_{ps}\hat{b}_{i}^\top = \mathbf{E}_{ps}x_{p}^i\top = x^{'i}_{p}.
        \end{equation}
    
    \item[(e)]Concatenate the encrypted vectors $b'_{i} (i \in {1, \dots, N})$ into an encrypted image $x'$.

\end{itemize}

Figure \ref{Encrypted Images} shows an example of images encrypted with this procedure, where block size $p$ was 16.

\subsection{Domain Adaptation}
As shown in Fig. \ref{Overview of DA}, a model $f_{\theta}$ pre-trained in the plain domain is transformed into a model $\hat{f}_{\theta}$ to reduce the influence of encryption prior to the fine-tuning of the pre-trained model by using a novel domain adaptation method. The procedure of the proposed adaptation is summarized here.

The domain adaptation is carried out in accordance with the embedding structure of ViT. $\mathbf{E_{bs}}$  in Eq. \ref{eq: ebs} is used to permutate blocks in an image, so it has a close relationship with position embedding $\mathbf{E_{pos}}$. $\mathbf{E_{pos}}$ includes the position information of a class token $x_{class}$. In contrast, $\mathbf{E_{bs}}$ does not consider the information of $x_{class}$. To fill the gap, $\mathbf{E_{bs}}$ is extended as 

\begin{equation}
  \mathbf{E_{bs}'} =
    \begin{bmatrix} 
          1 & 0 & 0 & \dots & 0 \\
          0 & k_{(1,1)} & k_{(1,2)} & \dots & k_{(1,N)} \\
          0 & k_{(2,1)} & k_{(2,2)} &\dots  & k_{(2,N)} \\
          \vdots &\vdots & \vdots & \ddots & \vdots \\
          0 & k_{(N,1)} & k_{(N,2)} & \dots & k_{(N,N)}\\
    \end{bmatrix},
\end{equation}
\begin{align*}
           \mathbf{E_{bs}'} &\in \mathbb{R}^{(N + 1) \times (N + 1)}.
\end{align*}

In the domain adaptation, $\mathbf{E_{pos}}$ is transformed as in

\begin{equation}\label{eq: pos}
    \mathbf{\hat{E}_{pos}} =\mathbf{E_{bs}'}\mathbf{E_{pos}}.
\end{equation}

By using the adaptation, the position embedding of the pre-trained model can be adapted to the position information permutated by the block scrambling of images. 
Similarly, patch embedding $\mathbf{E}$ is transformed to adapt the position information of every pixel randomly replaced by pixel shuffling, as in

\begin{equation}\label{eq: patch}
    \mathbf{\hat{E}} =\mathbf{E_{ps}}\mathbf{E}.
\end{equation}

From Eqs. \ref{eq: bs}, \ref{eq: ps}, \ref{eq: pos}, and \ref{eq: patch}, under the use of image encryption and domain adaptation, an adapted sequence of embedded patches is given by

\begin{align} \label{eq: enc input}
    \hat{z}_{0} =& \mathbf{E_{bs}'}[x_{class}; \mathbf{E_{ps}}x_{p}^1\mathbf{E_{ps}}\mathbf{E} ; \dots \mathbf{E_{ps}}x_{p}^N\mathbf{E_{ps}}\mathbf{E}]^\top + \mathbf{E_{bs}'}\mathbf{E_{pos}}\\ \notag
    =& \mathbf{E_{bs}'}[x_{class}; \hat{x}_{p}^1\mathbf{\hat{E}}; \dots \hat{x}_{p}^N\mathbf{\hat{E}}]^\top + \mathbf{E_{bs}'}\mathbf{E_{pos}}.
\end{align}\\
Finally, fine-tuning with encrypted images is applied to the pre-trained model corrected with the proposed adaptation.

\begin{table}[tb]
 \caption{Classification accuracy (\%) with and without \\ proposed method (CIFAR-10)}
 \label{table:cifar10}
 \centering
  \begin{tabular}{cc|c}
   \hline
    & \multicolumn{2}{c}{Domain Adaptation} \\
    \hline
    Encryption & Yes & No \\
    \hline
   Pixel shuffling & \textbf{99.05} & 97.78 \\
   Block scrambling & \textbf{98.93} & 94.25\\
   Pixel shuffling + Block scrambling & \textbf{98.98} & 72.86\\
   \hline
   Plain & \multicolumn{2}{c}{99.00}\\
   \hline
  \end{tabular}
\end{table}

\begin{figure*}[tb]
\centering
    \begin{tabular}{cc}
        \vspace{-60px}
        \hspace{10px}Training & \hspace{210px} Testing \\
        \hspace{-120px}
        \begin{minipage}{4truecm}
            \centering
              \includegraphics[bb=0 0 1140 860,scale=0.275]{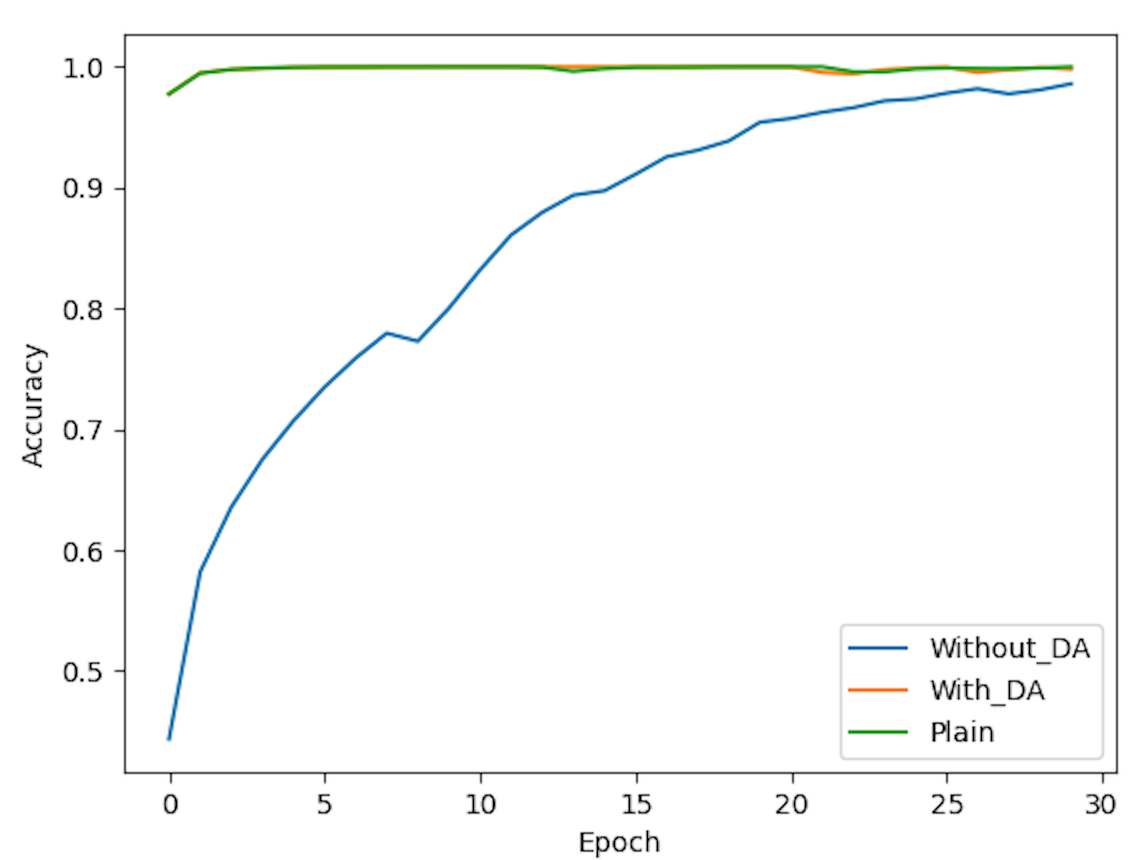}
        \end{minipage}
        &  \hspace{80px} 
        \begin{minipage}{4truecm}
            \centering
            \includegraphics[bb=0 0 1140 860,scale=0.275]{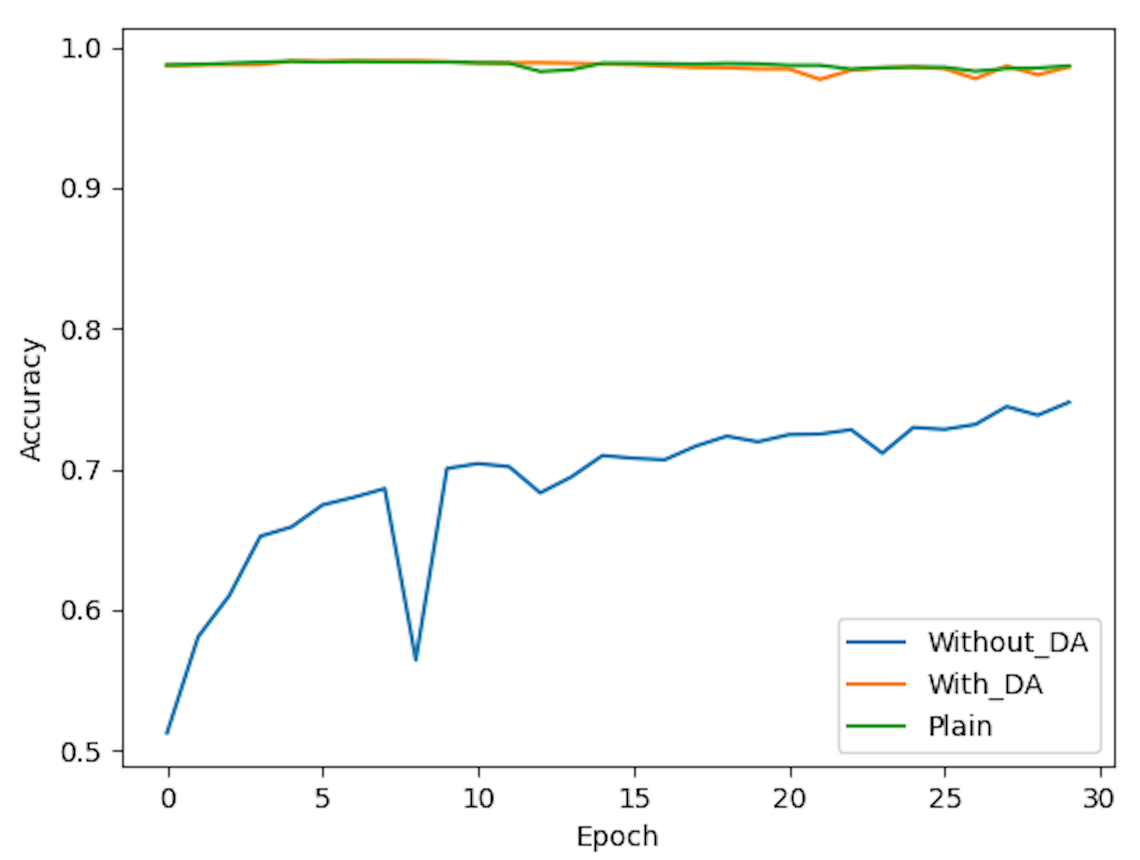}
            \end{minipage}
        \vspace{-40px}
        \\
        \hspace{-120px}
        \begin{minipage}{4truecm}
            \centering
              \includegraphics[bb=0 0 1140 860,scale=0.275]{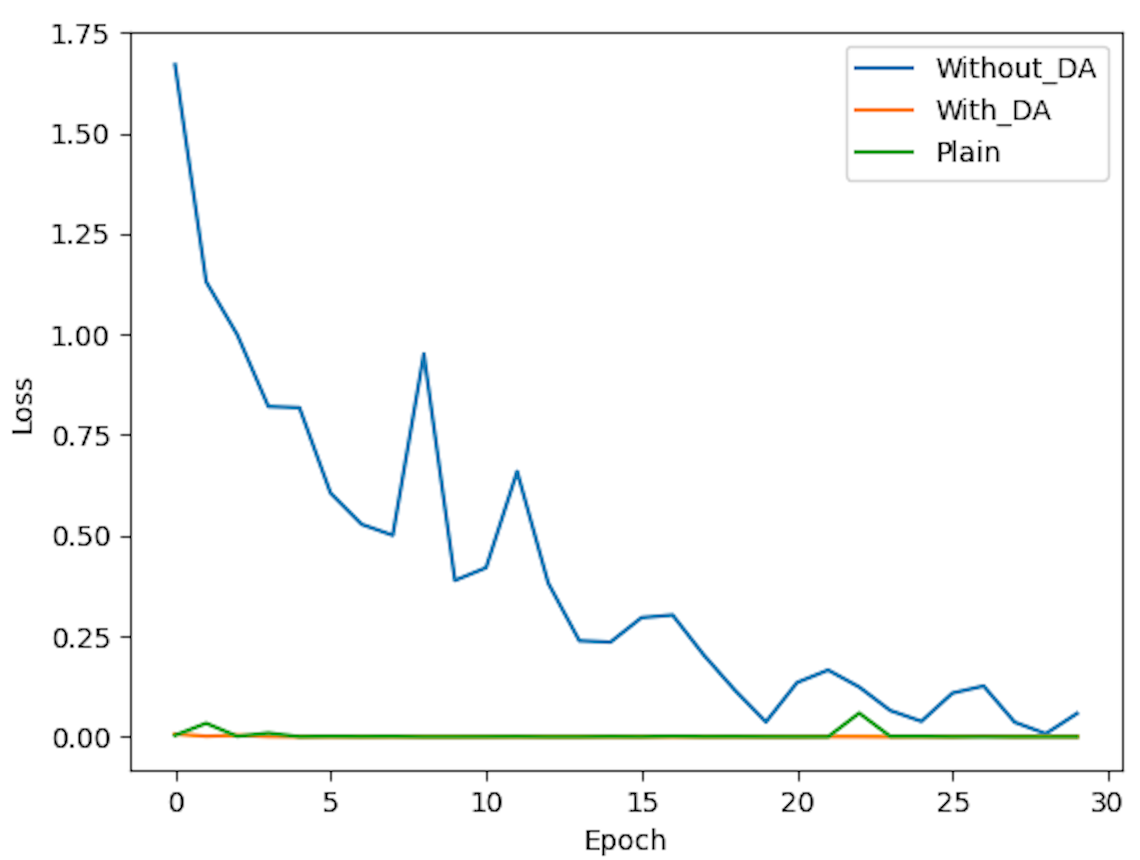}
        \end{minipage}
        & \hspace{80px}
        \begin{minipage}{4truecm}
            \centering
             \includegraphics[bb=0 0 1140 860,scale=0.275]{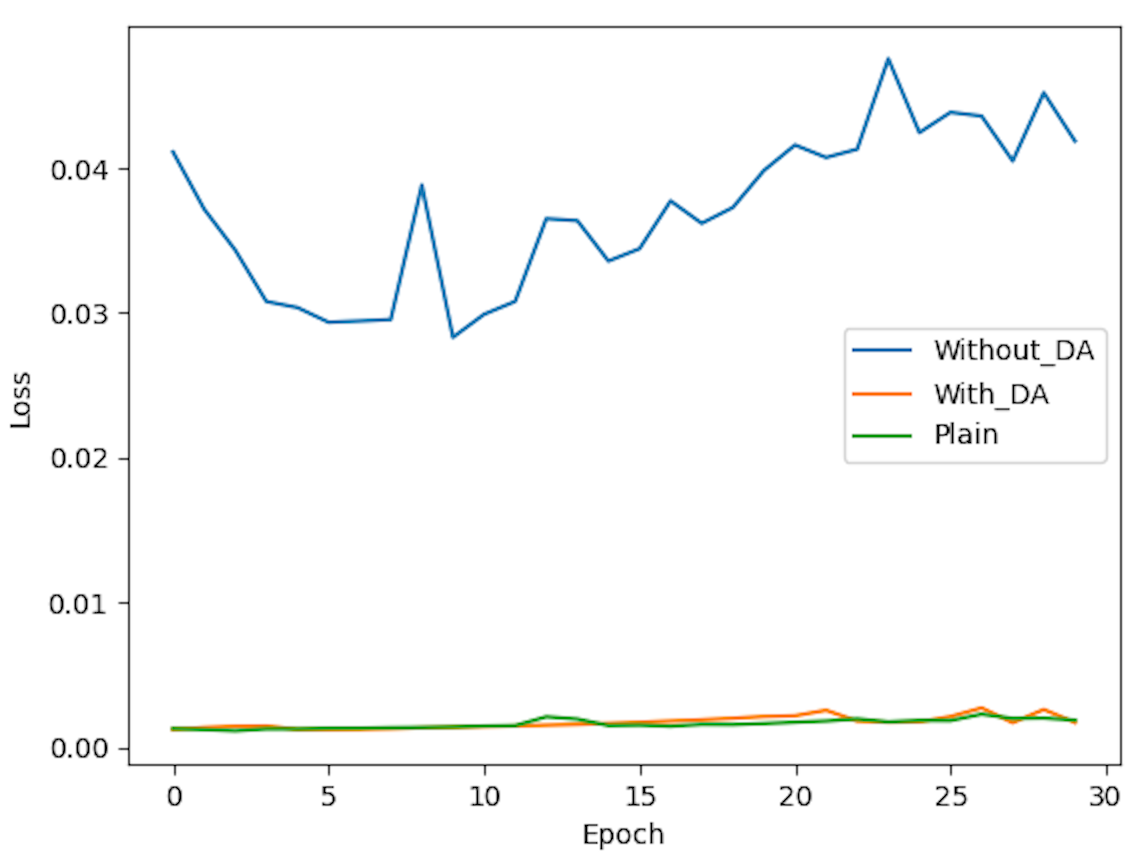}
        \end{minipage}\\
    \end{tabular}
    \caption{Efficiency of model training \\
(Top: classification accuracy per epoch, Bottom: loss value per epoch)}
    \label{Training efficiency}
\end{figure*}

\section{Experiments}
\subsection{Setup}
We conducted image classification experiments on the CIFAR-10 \cite{cifar10} and CIFAR-100 datasets \cite{cifar10}. CIFAR-10 consists of 60000 images with 10 classes (6000 images for each class), and CIFAR-100 consists of 60000 images with 100 classes (600 images for each class), where 50000 images were used for training and 10000 images were used for testing. Also, we fine-tuned a ViT-B\underline{ }16 model pre-trained with the ImageNet21k dataset, so we resized the image size from $32 \times 32 \times 3$ to $224 \times 224 \times 3$. Training parameters were a batch size of 32, a learning rate of 0.001, a momentum of 0.9, and a weight decay of 0.0005 on the stochastic gradient descent (SGD) for 15 epochs. A cross-entropy loss function was used as the loss function. The block size of the encryption was set to 16 to match the ViT patch size.

\begin{table*}[tb]
 \caption{Comparison with existing methods}
 \label{table:comparison}
 \centering
  \begin{tabular}{cccc}
   \hline
    \multirow{2}{*}{Encryption} & \multirow{2}{*}{Model} & \multicolumn{2}{c}{Accuracy (\%)} \\
        &    &  CIFAR-10 & CIFAR-100 \\
    \hline
   LE \cite{LE} & AdaptNet $+$ Shakedrop & 94.49 & 75.48 \\
   ELE \cite{madono2020} &  AdaptNet $+$ Shakedrop & 83.06 & 62.97 \\
   PE \cite{Pixel-Based} & ResNet-18 & 92.03 & - \\
   \multirow{2}{*}{EtC \cite{maung_privacy}} & ViT-B\underline{ }16 & 87.89 & -\\
    &  ConvMixer-256/8 & 92.72 & -\\
   Color-NeuraCrypt \cite{qi2023colorneuracrypt} & ViT-B\underline{ }16 & 96.20 & -\\
   Block scrambling $+$ Pixel position shuffling \cite{qi-san} & ViT-B\underline{ }16 & 96.64 & 84.42\\
   Block scrambling $+$ Pixel shuffling $+$ Domain adaptation (Proposed) & ViT-B\underline{ }16 & \textbf{98.98} & \textbf{92.58}\\
   \hline
   Plain & ShakeDrop & 96.70 & 83.59\\
   Plain & ResNet-18 & 95.53 & -\\
   Plain & ViT-B\underline{ }16 & 99.00 & 92.60\\
   Plain & ConvMixer-256/8 & 96.07 & -\\
   \hline
  \end{tabular}
\end{table*}

\subsection{Effects of Domain Adaptation}
\subsubsection{Classification Performance}
Table \ref{table:cifar10} shows the classification accuracy of models trained with encrypted images.  When training models without the proposed domain adaptation, the accuracy decreased compared with that of using plain images. In particular, when applying the combined use of pixel shuffling and block scrambling, the classification accuracy decreased significantly. In contrast, when training models with the domain adaptation, the accuracy was almost the same as that of using plain images. Accordingly, we confirmed that the proposed domain adaptation was effective in fine-tuning pre-trained models with encrypted images.

\subsubsection{Training Efficiency}
Figure \ref{Training efficiency} shows the classification accuracy and loss value per epoch during training and testing where the number of epochs was increased to 30. \par
For training, when using the domain adaptation during training, models fine-tuned with encrypted images (with\underline{ }DA) achieved almost the same training efficiency as those trained with plain images. In contrast, models fine-tuned without the domain adaptation (without\underline{ }DA) needed the larger number of epochs to achieve the same performance as plain models.  For testing, models with the domain adaptation also achieved the same performance as those with plain images, but models without the domain adaptation were overtrained.\par
Accordingly, the domain adaptation allows us not only to train a high-performance model without increasing training time even when using encrypted images, but to also avoid overtraining during test time.

\subsection{Comparison with Conventional Methods}
In Table \ref{table:comparison}, the proposed method is compared with the state-of-the-art methods for privacy preserving image classification with encrypted images such as LE \cite{LE}, PE \cite{Pixel-Based}, and ELE \cite{madono2020}.
From the table, the proposed method not only has the lowest performance degradation on both CIFAR-10 and CIFAR-100 datasets but also outperforms all existing methods in terms of the accuracy of image classification.

\section{Conclusion}
In this paper, we proposed a domain adaptation method to reduce the influence of image encryption for ViT models. The method is applied to a model pre-trained in the plain domain prior to the fine-tuning of the pre-trained model. In experiments, the proposed adaptation was demonstrated not only to reduce the performance degradation of models but to also achieve the highest classification accuracy in conventional methods on the CIFAR-10 and CIFAR-100 datasets. 

\section*{Acknowledgment}
This study was partially supported by JSPS KAKENHI (Grant Number JP21H01327) and JST CREST (Grant Number JPMJCR20D3).

\printbibliography

@ARTICLE{maung_privacy,
  author={AprilPyone, MaungMaung and Kiya, Hitoshi},
  journal={IEEE MultiMedia}, 
  title={Privacy-Preserving Image Classification Using an Isotropic Network}, 
  year={2022},
  volume={29},
  number={2},
  pages={23-33},
  doi={10.1109/MMUL.2022.3168441}}

@article{kiya2022overview,
year = {2022},
volume = {11},
journal = {APSIPA Transactions on Signal and Information Processing},
title = {An Overview of Compressible and Learnable Image Transformation with Secret Key and its Applications},
doi = {10.1561/116.00000048},
issn = {},
number = {1, e11},
author = {Kiya, Hitoshi and AprilPyone, MaungMaung and Kinoshita, Yuma and Imaizumi,Shoko and Shiota, Sayaka}
}

@inproceedings{ViT,
  author    = {Alexey Dosovitskiy and
               Lucas Beyer and
               Alexander Kolesnikov and
               Dirk Weissenborn and
               Xiaohua Zhai and
               Thomas Unterthiner and
               Mostafa Dehghani and
               Matthias Minderer and
               Georg Heigold and
               Sylvain Gelly and
               Jakob Uszkoreit and
               Neil Houlsby},
  title     = {An Image is Worth 16x16 Words: Transformers for Image Recognition
               at Scale},
  booktitle = {9th International Conference on Learning Representations, {ICLR} 2021,
               Virtual Event, Austria, May 3-7},
  publisher = {OpenReview.net},
  year      = {2021},
}

@Article{jimaging8090233,
AUTHOR = {Kiya, Hitoshi and Nagamori, Teru and Imaizumi, Shoko and Shiota, Sayaka},
TITLE = {Privacy-Preserving Semantic Segmentation Using Vision Transformer},
JOURNAL = {Journal of Imaging},
VOLUME = {8},
YEAR = {2022},
NUMBER = {9},
ARTICLE-NUMBER = {233},
URL = {https://www.mdpi.com/2313-433X/8/9/233},
PubMedID = {36135399},
ISSN = {2313-433X},
DOI = {10.3390/jimaging8090233}
}

@INPROCEEDINGS{LE,  author={Tanaka, Masayuki},  booktitle={2018 IEEE International Conference on Consumer Electronics-Taiwan (ICCE-TW)},   title={Learnable Image Encryption},   year={2018},  volume={},  number={},  pages={1-2},  doi={10.1109/ICCE-China.2018.8448772}}

@INPROCEEDINGS{madono2020,
  title= {Block-wise Scrambled Image Recognition Using Adaptation Network},
  author = {Madono, Koki and Tanaka, Masayuki and Onishi Masaki and Ogawa Tetsuji},
  booktitle = {Workshop on Artificial Intelligence of Things (AAAI WS)},year = {2020}
}

@ARTICLE{Pixel-Based,  author={Sirichotedumrong, Warit and Kinoshita, Yuma and Kiya, Hitoshi},  journal={IEEE Access},   title={Pixel-Based Image Encryption Without Key Management for Privacy-Preserving Deep Neural Networks},   year={2019},  volume={7},  number={},  pages={177844-177855},  doi={10.1109/ACCESS.2019.2959017}}

@ARTICLE{Encryption-Then-Compression,  author={Chuman, Tatsuya and Sirichotedumrong, Warit and Kiya, Hitoshi},  journal={IEEE Transactions on Information Forensics and Security},   title={Encryption-Then-Compression Systems Using Grayscale-Based Image Encryption for JPEG Images},   year={2019},  volume={14},  number={6},  pages={1515-1525},  doi={10.1109/TIFS.2018.2881677}}

@INPROCEEDINGS{F,
  author={Sirichotedumrong, Warit and Chuman, Tatsuya and Imaizumi, Shoko and Kiya, Hitoshi},
  booktitle={2018 IEEE International Conference on Multimedia and Expo (ICME)}, 
  title={Grayscale-Based Block Scrambling Image Encryption for Social Networking Services}, 
  year={2018},
  volume={},
  number={},
  pages={1-6},
  doi={10.1109/ICME.2018.8486525}}

@ARTICLE{Ito_access,
  author={Ito, Hiroki and Kinoshita, Yuma and Aprilpyone, Maungmaung and Kiya, Hitoshi},
  journal={IEEE Access},
  title={Image to Perturbation: An Image Transformation Network for Generating Visually Protected Images for Privacy-Preserving Deep Neural Networks},
  year={2021},
  volume={9},
  number={},
  pages={64629-64638},
  doi={10.1109/ACCESS.2021.3074968}}

@article{qi-san,
  author = {Qi, Zheng and AprilPyone, MaungMaung and Kinoshita, Yuma and Kiya, Hitoshi},
  title = {Privacy-Preserving Image Classification Using Vision Transformer},
  journal = {EURASIP European Signal Processing Conference, Belgrade, Serbia, August 31},
  year = {2022},
  pages={543-547},

}

@article{E,
  title={Unitary Transform-Based Template Protection and Its Application to l2-norm Minimization Problems},
  author={Nakamura, Ibuki and Tonomura,Yoshihide and Kiya, Hitoshi},
  journal={IEICE Transactions on Information and Systems},
  volume={E99.D},
  number={1},
  pages={60-68},
  year={2016},
  doi={10.1587/transinf.2015MUP0007}
}

@article{FL,
  url = {https://arxiv.org/abs/1610.05492},
  author = {Konečný, Jakub and McMahan, H. Brendan and Yu, Felix X. and Richtárik, Peter and Suresh, Ananda Theertha and Bacon, Dave},
  title = {Federated Learning: Strategies for Improving Communication Efficiency},
  journal = {arXiv},
  year = {2016},
}

@inproceedings{cifar10,
    author={Krizhevsky, Alex},
    title = {Learning multiple layers of features from tiny images},
    publisher={University of Toronto, Tech. Rep.}, 
    booktitle={}, 
    year={2009}
}

@misc{qi2023colorneuracrypt,
      title={Color-NeuraCrypt: Privacy-Preserving Color-Image Classification Using Extended Random Neural Networks}, 
      author={Qi, Zheng and AprilPyone, MaungMaung and Kiya, Hitoshi},
      year={2023},
      eprint={2301.04875},
      archivePrefix={arXiv},
}

@ARTICLE{Homomorphic1,
  author={Phong, Le Trieu and Aono, Yoshinori and Hayashi, Takuya and Wang, Lihua and Moriai, Shiho},
  journal={IEEE Transactions on Information Forensics and Security}, 
  title={Privacy-Preserving Deep Learning via Additively Homomorphic Encryption}, 
  year={2018},
  volume={13},
  number={5},
  pages={1333-1345},
  doi={10.1109/TIFS.2017.2787987}}

@INPROCEEDINGS{Homomorphic2,
  author={Wang, Yizhi and Lin, Jun and Wang, Zhongfeng},
  booktitle={2018 IEEE International Symposium on Circuits and Systems (ISCAS)}, 
  title={An Efficient Convolution Core Architecture for Privacy-Preserving Deep Learning}, 
  year={2018},
  volume={},
  number={},
  pages={1-5},
  doi={10.1109/ISCAS.2018.8350963}}

@misc{yala2021neuracrypt,
      title={NeuraCrypt: Hiding Private Health Data via Random Neural Networks for Public Training}, 
      author={Adam Yala and Homa Esfahanizadeh and Rafael G. L. D' Oliveira and Ken R. Duffy and Manya Ghobadi and Tommi S. Jaakkola and Vinod Vaikuntanathan and Regina Barzilay and Muriel Medard},
      year={2021},
      eprint={2106.02484},
      archivePrefix={arXiv},
}

@misc{chang2020attacks,
      title={Attacks on Image Encryption Schemes for Privacy-Preserving Deep Neural Networks}, 
      author={Alex Habeen Chang and Benjamin M. Case},
      year={2020},
      eprint={2004.13263},
      archivePrefix={arXiv},
      primaryClass={cs.CR}
}

@ARTICLE{maung_AD,
  author={AprilPyone, Maungmaung and Kiya, Hitoshi},
  journal={IEEE Transactions on Information Forensics and Security}, 
  title={Block-Wise Image Transformation With Secret Key for Adversarially Robust Defense}, 
  year={2021},
  volume={16},
  number={},
  pages={2709-2723},
  doi={10.1109/TIFS.2021.3062977}}

@article{KIYA20232022,
  title={Image and Model Transformation with Secret Key for Vision Transformer},
  author={Hitoshi Kiya and Ryota Iijima and AprilPyone, Maungmaung and Yuma Kinoshita},
  journal={IEICE Transactions on Information and Systems},
  volume={E106.D},
  number={1},
  pages={2-11},
  year={2023},
  doi={10.1587/transinf.2022MUI0001}
}

\end{document}